\pdfoutput=1
\documentclass[twocolumn]{svjour3}          % twocolumn
\smartqed  % flush right qed marks, e.g. at end of proof
\usepackage{graphicx}
\graphicspath{{./}}			
\DeclareGraphicsExtensions{.pdf,.jpeg,.png}
\usepackage{color}
\usepackage{subfig}
\usepackage{mathptmx}   
\usepackage{amssymb}                        
\usepackage{amsmath}                            
\usepackage{amsfonts}
\usepackage{mathtools}
\usepackage{algpseudocode}
\usepackage{algorithm}
\usepackage{tabularx}
\usepackage{booktabs}
\usepackage{colortbl} 
\usepackage{xcolor} 
\usepackage{xfrac}
\usepackage{multirow}
\usepackage{bm}
\usepackage{amsbsy}
\usepackage{fixmath}
\usepackage{lettrine}

\usepackage[numbers,square,sort&compress]{natbib}

\begin{document}\sloppy

%\begingroup
%\huge
%\centerline{\textbf{Title Page}} 
%\vspace{2cm} 
%
%\large
%Names of the authors: Ziheng Wang$^1$,  A. Majewicz Fey$^2$
%\vspace{1.5cm}
%\\
%Title: Deep Learning with Convolutional Neural Network for Objective Skill Evaluation in Robot-assisted Surgery\\
%\vspace{1cm}
%\\
%Affiliation of Authors: \\
%$^1$Department of Electrical Engineering, University of Texas at Dallas, Richardson, TX, 75080 USA \\
%\\
%              \vspace{0.25cm}
%              \\
%$^2$Department of Mechanical Engineering, University of Texas at Dallas, Richardson, TX, 75080 USA, and Department of Surgery at UT Southwestern Medical Center, Dallas, TX, 75390 \\
%\vspace{1cm}
%\\
%Corresponding Author:\\
%Ziheng Wang\\
%   \email{zihengwang@utdallas.edu} 
%\endgroup
%
%%\title{
%%Deep Learning with Convolutional Neural Network for Surgical Robot Kinematics Decoding and Skill Evaluation}

\title{
Deep Learning with Convolutional Neural Network for Objective Skill Evaluation in Robot-assisted Surgery}

\author{Ziheng Wang         \and
        Ann Majewicz Fey 
}

\institute{Ziheng Wang \at
              Department of Mechanical Engineering, The University of Texas at Dallas, Richardson, TX 75080, USA. \\
              \email{zihengwang@utdallas.edu}          
           \and
           Ann Majewicz Fey \at
              Department of Mechanical Engineering, The University of Texas at Dallas, Richardson, TX 75080, USA. \\
              Department of Surgery, UT Southwestern Medical Center, Dallas, TX 75390, USA.
}

\date{Received: 10 January 2018 / Accepted: 11 September 2018}

\maketitle

%%%%%%%%%%%%%%%%%%%%%%%%%%%%%%%%%%%%%%%%%%%%%%%%%%%%%%%%%%%%%%%%%%%%%%%%%%%%%%%%
\begin{abstract} \hfill \break

\noindent \textit{Purpose:} 
With the advent of robot-assisted surgery, the role of data-driven approaches to integrate statistics and machine learning is growing rapidly with prominent interests in objective surgical skill assessment. 
However, most existing work requires translating robot motion kinematics into intermediate features or gesture segments that are expensive to extract, lack efficiency, and require significant domain-specific knowledge.

\noindent \textit{Methods:} 
We propose an analytical deep learning framework for skill assessment in surgical training. A deep convolutional neural network is implemented to map multivariate time series data of the motion kinematics to individual skill levels.

\noindent \textit{Results:} 
We perform experiments on the public minimally invasive surgical robotic dataset, JHU-ISI Gesture and Skill Assessment Working Set (JIGSAWS).
Our proposed learning model achieved a competitive accuracy of 92.5\%, 95.4\%, and 91.3\%, in the standard training tasks: \textit{Suturing}, \textit{Needle-passing}, and \textit{Knot-tying}, respectively. Without the need of engineered features or carefully-tuned gesture segmentation, our model can successfully decode skill information from raw motion profiles via end-to-end learning. 
Meanwhile, the proposed model is able to reliably interpret skills within 1-3 second window, without needing an observation of entire training trial.

\noindent \textit{Conclusion:} 
This study highlights the potentials of deep architectures for an proficient online skill assessment in modern surgical training.

\keywords{
Surgical robotics, surgical skill evaluation, motion analysis, deep learning, convolutional neural network}

\end{abstract}

%%%%%%%%%%%%%%%%%%%%%%%%%%%%%%%%%%%%%%%%%%%%%%%%%%%%%%%%%%%%%%%%%%%%%%%%%%%%%%%%
\section{INTRODUCTION}
Due to the prominent demand for both quality and safety in surgery, it is essential for surgeon trainees to achieve required proficiency levels before operating on patients~\cite{roberts2006evolution}. 
An absence of adequate training can significantly compromise the clinical outcome, which has been shown in numerous studies~\cite{reznick2006teaching,aggarwal2010training,birkmeyer2013surgical}. 
Effective training and reliable methods to assess surgical skills are thus critical in supporting trainees in technical skill acquisition~\cite{darzi2001assessment}.
Simultaneously, current surgical training is undergoing significant changes with a rapid uptake of minimally invasive robot-assisted surgery. 
However, despite advances of surgical technology, most assessments of trainee skills are still performed via outcome-based analysis~\cite{bridgewater2003surgeon}, structured checklists, and rating scales~\cite{goh2012gears, aghazadeh2015GEARS,niitsu2013OSAT}. Such assessment requires large amounts of expert monitoring and manual ratings, and can be inconsistent due to biases in human interpretations~\cite{reiley2011objective}. Considering the increasing attention to the efficiency and effectiveness of assessment and targeted feedback, conventional methods are no longer adequate in advanced surgery settings~\cite{vedula2017objective}.

%Current surgical training is undergoing significant changes with a rapid uptake of robot-assisted technology. Such advance of surgical systems enables a massive influx of sensory data from surgical robots or simulators~\cite{moustris2011robotsurgery}.
Modern robot-assisted surgical systems are able to collect a large amount of sensory data from surgical robots or simulators~\cite{moustris2011robotsurgery}. This high volume data could reveal valuable information related to the skills and proficiencies of the operator.  
However, analyzing such complex surgical data can be challenging.
Specifically, surgical motion profiles by nature are nonlinear, non-stationary stochastic processes~\cite{cheng2015time,klonowski2009everything} with large variability, both throughout a procedure, as well within repetitions of the same type of surgical task (e.g., suture throws)~\cite{reiley2009task}. 
In addition, the high dimensionality of the data creates an additional challenge for accurate and robust skill assessments~\cite{reiley2011objective}. Further, although several surgical assessment methods have been developed, methods to autonomously coach the trainee are lacking.
%implementation of automated coaching requires efficient techniques to inform how training curriculum can evolve to appropriately maximize training proficiency.   
Towards this aim, there is a great need to develop techniques for quicker and more effective surgical skill acquisition~\cite{kassahun2016surgical,vedula2017objective}.
In this paper, we are particularly interested in online skill assessment methods that could pave the way for autonomous surgical coaching.

\vspace{-10pt}
\subsection{Previous Approaches in Objective Skill Assessment}
Different objective skill assessment techniques have been reported in the literature~\cite{kassahun2016surgical}. Current approaches with a focus on surgical motions can be divided into two main categories: 1) descriptive statistic analysis, and 2) predictive modeling-based methods.
Descriptive statistic analysis aims to compute features from motion observations to quantitively describe skill levels. Specifically, summary features, such as movement time~\cite{judkins2009objective,liang2017motion,trejos2014force}, path length~\cite{judkins2009objective}, motion jerk~\cite{liang2017motion}, curvature~\cite{judkins2009objective}, etc., are widely used and have shown to have high correlations with surgical skills. Other novel measures of motion, such as energy expenditure~\cite{poursartip2017analysis}, semantic labels~\cite{ershad2016meaningful}, tool orientation~\cite{sharon2017ori_basedmetrics}, force~\cite{trejos2014force}, etc., can also provide discriminative information in measuring skills.
However, this approach involves manual feature engineering, requires task-specific knowledge and significant effort to design optimal skill metrics~\cite{shackelford2017metrics}. In fact, defining the best metrics to capture adequate information and be generalized enough to apply across different types of surgery or groups of surgeons remains an open problem~\cite{judkins2009objective,kassahun2016surgical,fard2018automated,stefanidis2009metrics}. 

In contrast to descriptive analysis, predictive modeling-based methods aim to predict surgical skills from motion data. This method can be further categorized into 1) descriptive, and 2) generative modeling. In descriptive modeling, models are learnt by transforming raw motion data to intermediate interpretations and summary features. Coupled with advanced feature selection, these pre-defined representations are subsequently fed into learning models as an input for skill assessment. In the literature, machine learning (ML) algorithms are commonly explored for modeling, such as k-nearest neighbors (kNN), logistic regression (LR), support vector machines (SVM), and linear discriminant analysis (LDA). Such algorithms yielded a skill predictive accuracy between 61.1\% and 95.2\%~\cite{chmarra2010skills,vedula2016taskseg,fard2018automated,poursartip2017energy}.
%Chmarra \emph{et al.} applied linear discriminant analysis (LDA)  using tool motion features to classify skills with an approximate accuracy of 74\%~\cite{chmarra2010skills}.
%In ~\cite{fard2018automated}, Fard \emph{et al.} explored the validity of k-nearest neighbors (kNN), logistic regression (LR), and support vector machines (SVM), based on motion summary features. Classifiers could obtain accuracies in the range of 63.0\% and 95.2\%. 
Forestier \emph{et al.} developed a novel vector space model (VSM) to assess skills via learning from the \textit{bag of word}, a collection of discretized local features (strings) obtained from motion data~\cite{forestier2017jigsaw}. In~\cite{brown2017using}, Brown \emph{et al.} explored an ensemble approach, which combines multiple ML algorithms for modeling, and was able to predict rating scores with moderate accuracies (51.7\% to 75.0\%).
More recently, Zia \emph{et al.} utilized nearest neighbor (NN) classifiers with a novel feature fusion (texture-, frequency- and entropy-based features) and further improved skill assessment with accuracy ranging from 99.7\% to 100\%~\cite{zia2018skill}. 
Although the descriptive modeling-based approaches show their validity in revealing skill patterns and underlying operation structures, the model accuracy and validity are typically limited by the quality of extracted features. Considering the complex nature of surgical motion profiles, critical information has the potential to be discarded within the feature extraction and selection process.
Alternatively, in generative modeling, temporal motion data are usually segmented into a series of predefined rudimentary gestures for certain surgical tasks. Using generative modeling algorithms, such as Hidden Markov Model (HMM) and its variants, several class-specific skill models were trained for each level and achieved accuracy ranging from 94.4\% to 100\%~\cite{reiley2009task,tao2012sparse}.
However, the segmentation of surgical gestures from surgeon motions can be a strenuous process. HMM models usually require large amounts of time and computational effort for parameter tuning and model development.
Further, one typical deficiency is that the skill assessment is obtained at the global task level, i.e., at the end of each operation. It requires an entire observation for each trial. This drawback potentially undermines the goal for an efficient online surgical skill assessment.

\vspace{-10pt}
\subsection{Proposed Approach}
Deep learning, also referred to as deep structured learning, is a set of learning methods that allow a machine to automatically process and learn from input data via hierarchical layers from low to high levels~\cite{2015natureDL,2015DL_review}. 
These algorithms perform feature self-learning to progressively discover abstract representations during the training process.   
Due to its superiority in complex pattern recognition, this approach dramatically improves the state of the art. Currently, deep learning models have achieved success in strategic games~\cite{silver2016alphaGO}, speech recognition~\cite{graves2013speech}, medical imaging~\cite{esteva2017dermatologist}, health informatics~\cite{Ng2017cardiologist}, and more.
In the study of robotic surgical training, DiPietro~\emph{et al.} first apply deep learning based on Recurrent Neural Networks for gesture and high-level task recognition~\cite{dipietro2016recognizing}. 
Still, relatively little work has been done to explore deep learning approaches for surgical skill assessment.

\begin{figure*}[tb]
      \centering
     \includegraphics [width=0.95\linewidth,clip,trim=5pt 20pt 0pt 35pt]{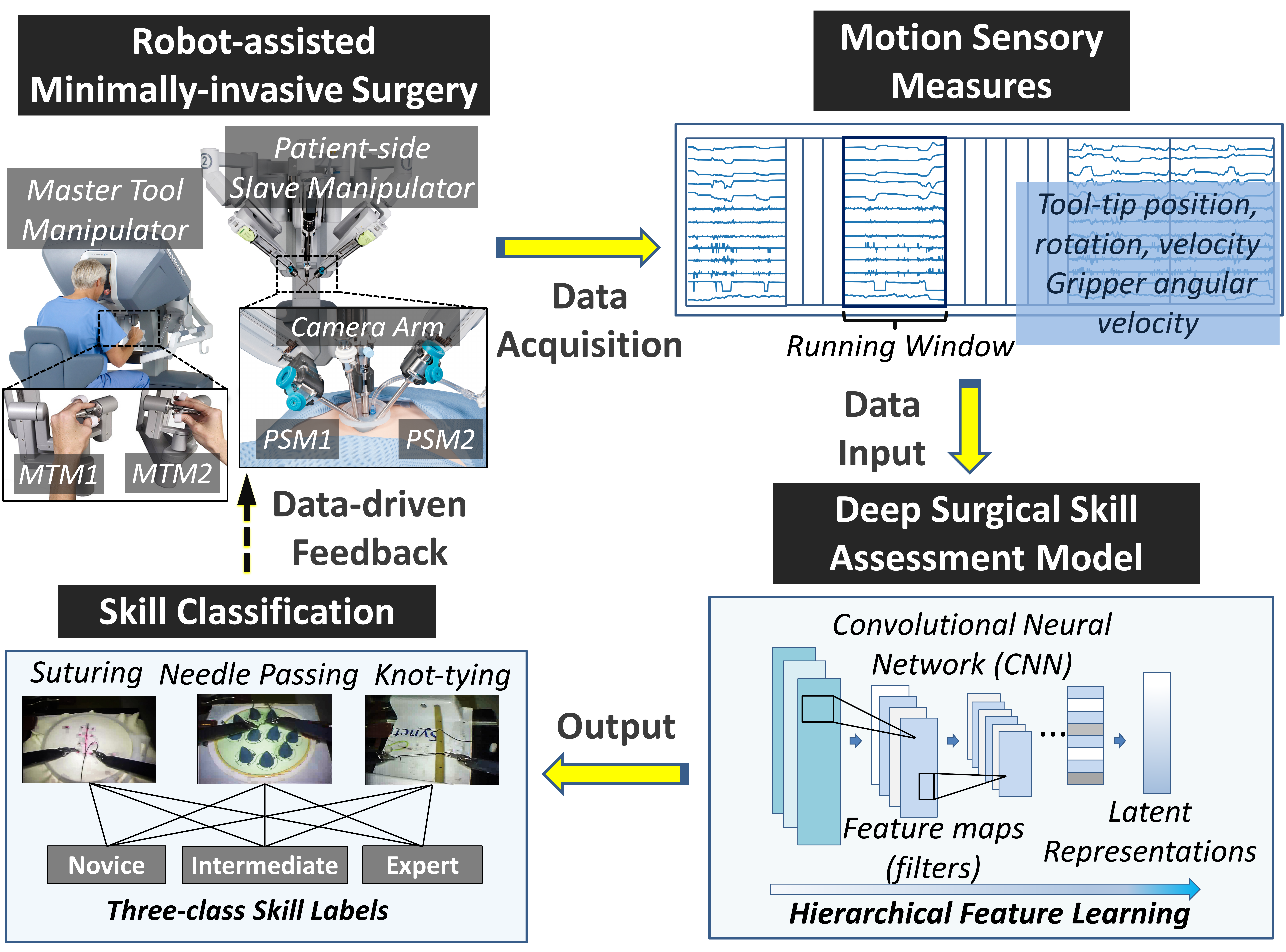}
     \caption{ {\bf An end-to-end framework for online skill assessment  in robot-assisted minimally-invasive surgery.} The framework utilizes window sequences of multivariate motion data as an input, recorded from robot end-effectors, and outputs a discriminative assessment of surgical skills via a deep learning architecture.} 
      \label{fig: fig1} 
\end{figure*}

In this paper, we introduce and evaluate the applicability of deep learning for a proficient surgical skill assessment. Specifically, a novel analytical framework with deep surgical skill model is proposed to directly process multivariate time series via an automatic learning. We hypothesize the learning-based approach could help to explore the intrinsic motion characteristics for decoding skills and promote an optimal performance in online skill assessment systems.
Fig~\ref{fig: fig1} shows the end-to-end pipeline framework. Without performing manual feature extraction and selection, latent feature learning is automatically employed on multivariate motion data and directly outputs classifications. To validate our approach, we conduct experiments on the public robotic surgery dataset, JIGSAW~\cite{gao2014JIGSAW}, in analysis of three independent training tasks: \textit{Suturing} (SU), \textit{Needle-passing} (NP), and \textit{Knot-tying} (KT).
To the best of our knowledge, it is the first study to employ a deep architecture for an objective surgical skill analysis. The main contributions of this paper can be summarized as:
\begin{enumerate}
	\item[--] An novel end-to-end analytical framework with deep learning for skill assessment based on high-level analysis of surgical motion. 
	\item[--] Experimental evaluation of our proposed deep skill model.
	\item[--] Application of data augmentation leveraging the limitation of small-scale JIGSAWS dataset, discussion on the effect of labeling approaches on the assessment accuracy, and exploration of validation schemes applicable for deep-learning-based development.
\end{enumerate}

In the remainder of this paper we first present our proposed approach and implementation details in Section~\ref{sec: model}.
We then conduct experiments on JIGSAW dataset to validate the model in Section~\ref{sec: experiments}. Data pre-processing, training, and evaluation approaches are given. Then, we present our results in Section~\ref{sec: results} and discussions in Section~\ref{sec: discussion}. Last, we conclude this paper in Section \ref{sec:conclusion}.

%%%%%%%%%%%%%%%%%%%%%%%%%%%%%%%%%%%%%%%%%%%%%%%%%%%%%%%%%%%%%%%%%%%%%%%%%%%%%%%%

\begin{figure*}[tb]
      \centering
     \includegraphics [width=1\linewidth,clip,trim=0pt 40pt 0pt 5pt]{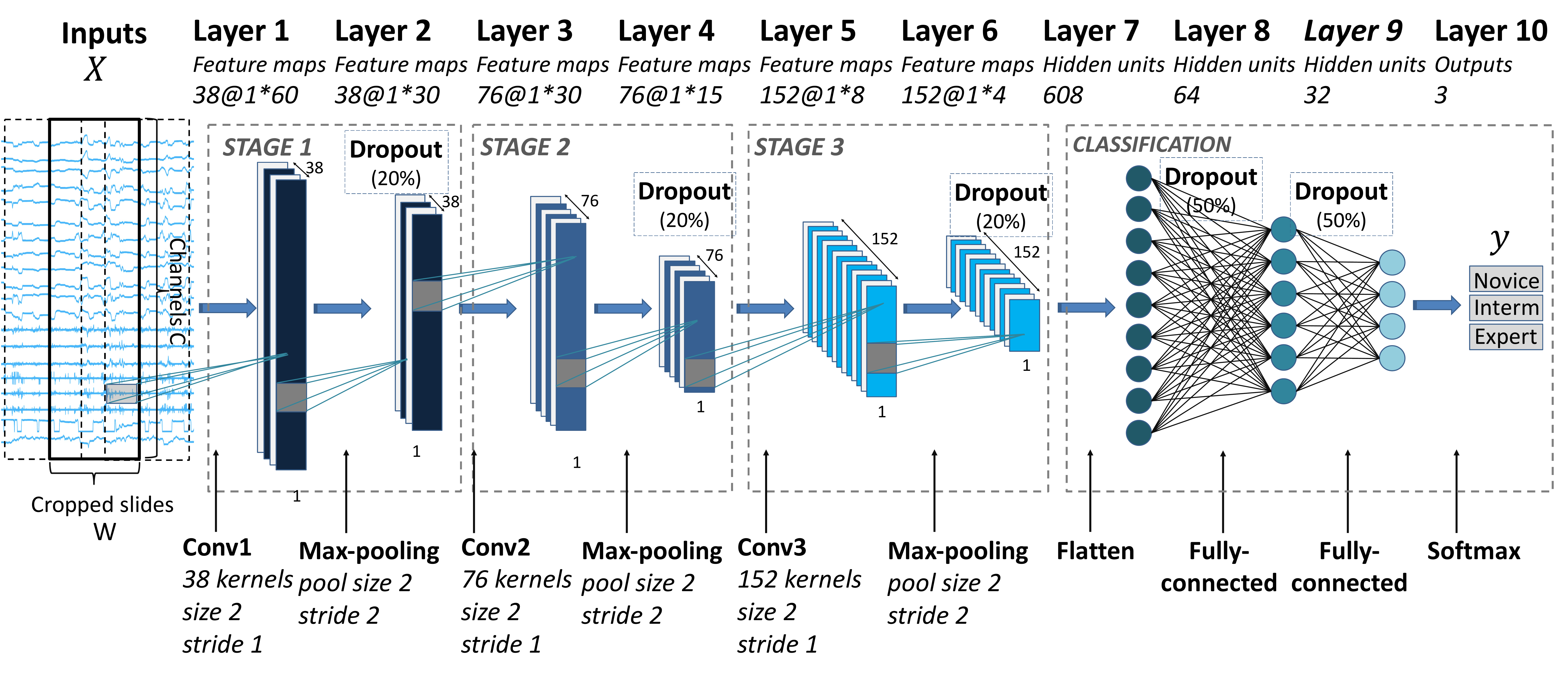}
     \caption{ {\bf Illustrations of the proposed deep architecture using a 10-layer convolutional neural network.} The window width $W$ used in this example is 60. Starting from the inputs, this model consists of three conv-pool stages with a convolution and max-pooling each, one flatten layer, two fully-connected layers, and one softmax layer for outputs. Note that the max-pooling dropout (with probability of 20\%) and fully-connected dropout (with probability of 50\%) is applied during training.}
      \label{fig: fig2}
\end{figure*}

\section{DEEP SURGICAL SKILL CLASSIFICATION MODEL}
\label{sec: model}

Our deep learning model for surgical skill assessment is motivated from studies in multiple domains~\cite{2015DL_review,langkvist2014review,gamboa2017deeptimeseries}. In this section, we introduce a deep architecture using Convolutional Neural Network (CNN) to assess surgical skills from an end-to-end classification. 

\subsection{Problem Formulation}
Here, the assessment of surgical skills is formalized as a supervised three-class classification problem, where the input is multivariate time series (MTS) of motion kinematics measured from surgical robot end-effectors, $X$, and the output is the predicted labels representing corresponding expertise levels of trainees, which can be one-hot encoded as $y \in \{1 :``Novice", 2:``Intermediate'', 3 :``Expert" \}$. Typically, ground-truth skill labels are acquired from expert ratings, crowdsourcing, or self-reporting experience.  
The objective cost function for training the network is defined as a multinomial cross-entropy cost, $J$, as shown in Eq.~\ref{Eq: cross-entropy}.

\begin{equation}
\label{Eq: cross-entropy}
J(\theta) = -\displaystyle\sum_{i=1}^{m}\displaystyle\sum_{k=1}^{K}{ 1 \{y^{(i)}=k \}  \log{p(y^{(i)}=k | x^{(i)}; \theta )}  }
\end{equation} 
where $m$ is the total number of training examples, $K$ is the class number, $K=3$, and $p(y^{(i)}=k | x^{(i)}; \theta)$ is the conditional likelihood that the predicted label $y^{(i)}$ on a single training example $x^{(i)}$ is assigned to class $k\in K$, given specific trained model parameters $\theta$.

\subsection{Model Architecture}
The architecture of the proposed neural network consists of five types of layers: convolutional layer, pooling layer, flatten layer, fully-connected layer and softmax layer. 
Fig.~\ref{fig: fig2} shows a 10-layer working architecture and parameter settings used in the network. Note that, the depth of the network is chosen after trial-and-error from the training/validation procedure.

The network takes the slide of length $W$ from $C$-channel sensory measurements as input, which is a $W\times C$ matrix, where $C$ is the number of channels, or dimensions, of input time series. 
Then, input samples are first processed by three convolution-pooling (Conv-pool) stages, where each stage consists of a convolution layer and a max-pooling layer. 
Each convolution layer has different numbers of kernels with the size of 2 and each kernel is convoluted with the input matrix of the layer with a stride of 1. Specifically, the first convolution (\textit{Conv1}) filters the $W\times 38$ input matrix with 38 kernels; the second convolution with 76 kernels (\textit{Conv2}) will filter the corresponding output matrix of previous layer; and the third convolutional layer (\textit{Conv3}) filters with 152 kernels. To reduce the dimensionality of the feature maps and avoid overfitting, corresponding connections of each convolution are followed by a max-pooling operation. The max-pooling operations take the output of convolution layer as input, and downsample the extracted feature maps, where each local input patch is replaced by taking the maximum value of each channel over the patch. The size of max-pooling is set as 2 with a stride of 2. 
In this network, we use the rectified linear unit (ReLU) as the activation function to add nonlinearity in all convolutional layers and fully-connected layers~\cite{nair2010relu}. Finally, we apply a softmax logistic regression to produce a distribution of probability over three classes for the output layer. 

\subsection{Implementation}
To implement the proposed architecture, the deep learning skill model is trained from scratch, which does not require any pre-trained model.The network algorithm is implemented using Keras library with Tensorflow backend based on Python 3.6~\cite{chollet2015keras}. We first initialize parameters at each layer using the Xavier initialization method~\cite{2015DL_review}, where biases are initialized as zeros, the weights at each layer are initialized from a Gaussian distribution with mean 0 and a variance of $1/N$, where $N$ specifies the number of neurons in the previous layer.

During the optimization, our network is trained end-to-end by minimizing the multinomial cross-entropy cost between the predicted and ground-truth labels, as defined in Eq.~\ref{eq1}, at the learning rate, $\epsilon$, of 0.0001. To train the net efficiently, we run mini-batch updates of gradient descent, which calculate network parameters on a subset of the training data at each iteration~\cite{li2014minibatch}. The size of mini batches is set to 600. A total of 300 epochs for training were run in this work. 
The network parameters are optimized by an Adam solver~\cite{kingma2014adam}, which computes adaptive learning rates for each neuron parameter via estimates of first and second moment of the gradients. The exponential decay rates of the first and second moment estimates are set to 0.9 and 0.999, respectively.  
Also, to achieve better generalization and model performance, we apply a stochastic dropout regularization to our neural network during training time. Components of outputs from specific layers of networks are randomly dropped out at a specific probability~\cite{srivastava2014dropout}. This method has proven its effectiveness to reduce over-fitting in complex deep learning models~\cite{wu2015maxpolldrop}. In this study, we implement two strategies of dropout: one is the max-pooling dropout on the layers of max-pooling after ReLU non-linearity; another regularization is the fully-connected dropout on the fully-connected layers. The probabilities of dropout for the max-pooling and fully-connected dropout are set at 0.2 and 0.5, respectively.  
As mentioned above, the hyper-parameters used for CNN implementation include the learning rate, mini-batch size, epoch, number of filters, stride and size of kernel, and dropout rates in the max-pooling and fully-connected layers. These hyper-parameters are chosen and fine-tuned by employing the validation set, which is split from training data. We save the best model, as evaluated on validation data, in order to obtain an optimal prediction performance. 

%%%%%%%%%%%%%%%%%%%%%%%%%%%%%%%%%%%%%%%%%%%%%%%%%%%%%%%%%%%%%%%%%%%%%%%%%%%%%%%%
\section{EXPERIMENT SETUP}
\label{sec: experiments}

\subsection{Dataset}
Our dataset comes from the JHU-ISI Gesture and Skill Assessment Working Set (JIGSAWS), the only public-available minimally invasive surgical database, which is collected from the \textit{da Vinci} tele-robotic surgical system~\cite{gao2014JIGSAW,ahmidi2017dataset}. 

The \textit{da Vinci} robot is comprised of two master tool manipulators (MTMs) on left and right sides, two patient-sides slave manipulators (PSMs), and an endoscopic camera arm. Robot motion data are captured (sampling frequency 30 Hz) as multivariate time series with 19 measurements for each end-effector: tool tip Cartesian positions ($x$, $y$, $z$), rotations (denoted by a $3 \times 3$ matrix $R$), linear velocities ($v_x$, $v_y$, $v_z$), angular velocities ($\omega_x'$, $\omega_y'$, $\omega_z'$), and the gripper angle $\theta$. Details of the JIGSAWS kinematic motion data are summarized in Table~\ref{tab: JIGSAW}. 

\begin{figure*}[tb]
      \centering
   \includegraphics [width=0.85\linewidth,clip,trim=0pt 0pt 0pt 0pt]{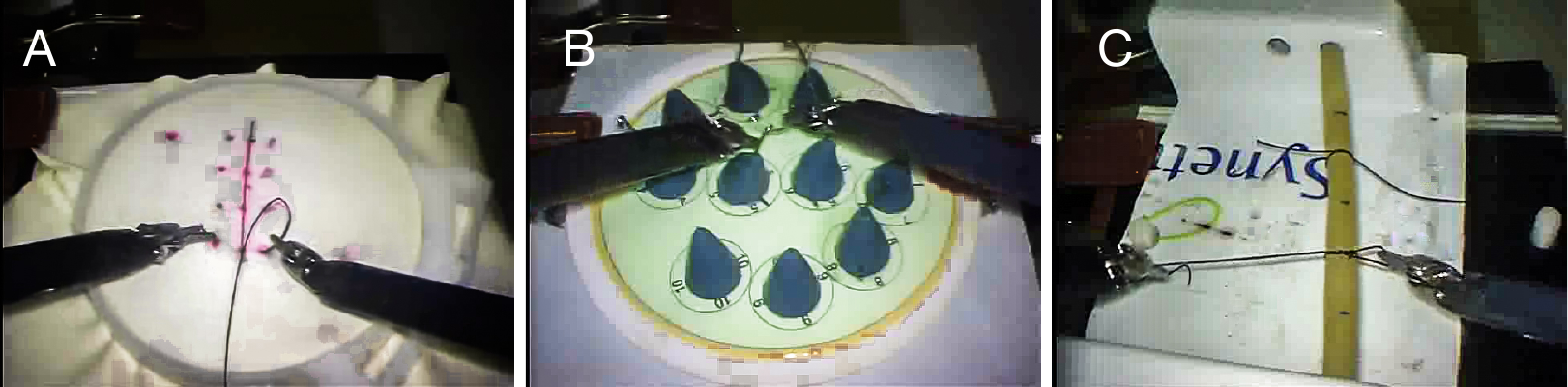}
    \caption{{\bf Snapshots of operation tasks during robot-assisted minimally invasive surgical training}. The operations are implemented using the \textit{da Vinci} robot and are reported in JIGSAWS~\cite{gao2014JIGSAW}: (A) \textit{Suturing}, (B) \textit{Needle-passing}, (C) \textit{Knot-tying}.}
      \label{fig: fig3}
\vspace{-0.3cm}
\end{figure*}

The dataset contains recordings from eight surgeons with varying robotic surgical experience. Each surgeon performed three different training tasks, namely, \textit{Suturing} (SU),  \textit{Knot-tying} (KT), and  \textit{Needle-passing} (NP),  and repeated each task five times. All three tasks are typically standard components in surgical skill training curricula~\cite{gao2014JIGSAW}. An illustration of the three operation tasks is shown in Fig.~\ref{fig: fig3}. 
The two ways in which skill labels are reported in JIGSAWS dataset are: (1) self-proclaimed skill labels based on practice hours with \textit{expert} reporting greater than 100 hours, \textit{intermediate} between 10-100 hours, and \textit{novice} reporting less than 10 hours of total surgical robotic operation time, and (2) a modified global rating scale (GRS) ranging from 6 and 30, manually graded by an experienced surgeon. 
In this study, we use the self-proclaimed skill levels and GRS-based skill levels as the ground-truth labels for each surgical trial, respectively. In order to label surgeons skill levels using GRS scores, inspired from~\cite{fard2018automated}, thresholds of 15 and 20 are used to divide surgeons into \textit{novice}, \textit{intermediate}, and \textit{expert}, in tasks of \textit{Needle-passing} and \textit{Knot-tying}, and thresholds of 19 and 24 are used in \textit{Suturing} for skill labeling.

\begin{table*}[tb]
\begin{centering}
\renewcommand\arraystretch{1.6}
\renewcommand\tabcolsep{11pt}
\footnotesize
\caption{{\bf Variables of sensory signals from end-effectors of \textit{da Vinci} robot.} These variables are captured as multivariate time series data in each surgical operation trial.}
\label{tab: JIGSAW}
\centering
\begin{tabular}{lcccc}
\toprule[\heavyrulewidth]
\multicolumn{2}{c}{ \bf{End-effector Category}} & \multicolumn{1}{c}{ \bf{Description} } & \bf{Variables} & \bf{Channels} \\
\midrule[\heavyrulewidth]    
\multirow{2}{*}{\parbox{4cm}{\bf{Master Tool \\ Manipulator (MTM)}}} & \textit{MTM1} & \multirow{2}{*}{\parbox{5cm}{Positions (3), rotation matrix (9), velocities (6) of tool tip, gripper angular velocity (1)}}  & \multirow{2}{*}{\parbox{2.5cm}{$x$, $y$, $z$, $R \in  {\rm I\!R}^{3\times 3}$, \\ $v_x$, $v_y$, $v_z$, $\omega_x'$, $\omega_y'$, $\omega_z'$, $\alpha$}} & \multirow{2}{*}{$19\times2$}\\
 & \textit{MTM2} &  &  &  \\
\midrule[\heavyrulewidth]    
\multirow{2}{*}{\parbox{4cm}{\bf{Patient-side\\Manipulator (PSM)}}} & \textit{PSM1} & \multirow{2}{*}{\parbox{5cm}{Positions (3), rotation matrix (9), velocities (6) of tool tip, gripper angular velocity (1)}} & \multirow{2}{*}{\parbox{2.5cm}{$x$, $y$, $z$, $R \in {\rm I\!R}^{3\times 3}$, \\ $v_x$, $v_y$, $v_z$, $\omega_x'$, $\omega_y'$, $\omega_z'$, $\alpha$}} & \multirow{2}{*}{$19\times2$} \\
 & \textit{PSM2} &  &  & \\
\bottomrule[\heavyrulewidth]
\end{tabular}
\end{centering}
\end{table*}

\subsection{Data Preparation \& Inputs}

\paragraph{$Z$-normalization} Due to differences in the scaling ranges and offset effects of the sensory data, the data fed into the neural network are first normalized with a $z$-normalization process. Each channel of raw data, $x$, is normalized individually as $z = \frac{x- \mu}{\sigma}$, where $\mu$ and $\alpha$ are the mean and standard deviation of vector $x$. This normalization process can be performed online by feeding the network with the batch of sensory data.

\paragraph{Data Augmentation} One challenge for developing a robust skill model with our approach comes from the lack of large-scale data samples in JIGSAWS, where the number of labeled samples is only 40 in total (8 subjects with 5 trial repetitions) for each surgical task. Generally, deep learning might suffer from overfitting if the size of available dataset is limited~\cite{2015natureDL}. To overcome this problem, data augmentation is introduced to prevent overfitting and improve generalization of the deep learning model.
This has been seen so far mostly in image recognition, where several methods, such as scaling, cropping, and rotating are used~\cite{krizhevsky2012imagenet,he2016deep}. Inspired from the computer vision community, similar augmentation techniques were applied for time series data to enlarge small-sized datasets and increase decoding accuracy \cite{cui2016MCNN,le2016dataaug,um2017data}.
In this study, to support the network in learning, we adapted the augmentation strategy and introduced a two-step augmentation process before inputting data into our network. 
First, followed by $z$-normalization, we viewed and separated the surgical motion data from master (MTMs) and patient-side manipulators (PSMs) as two distinct sample instances, while the class labels for each trial were preserved. This procedure is also appropriate in cases where the MTMs and PSMs are not directly correlated (e.g. position scaling, or other differences in robot control terms). 
Then, we carried out a label-preserving cropping with a sliding window, where the motion sub-sequences were extracted using crops, i.e., sliding a fixed-size window within the trial. The annotation for each window is identical to the class label of original trial, from which the sub-sequences are extracted. One advantage of this approach is that it leads to larger-scale sets for the robust training and testing of the network. Also, this technique allows us to format time series as equal-length inputs, regardless of the varied lengths of original data. 
The pseudo-code of sliding-window cropping algorithm is shown in Algorithm~\ref{alg: cropwindow}, where $X$ is the input motion data, $s$ is the output crops (i.e., sub-sequences), $W$ is the sliding window size and $L$ is the step size. After experimenting based on trial-and-error, we chose a window size $W=60$ and a step size  $L=30$ in this work. 
Overall, by applying the aforementioned data augmentation process on the original dataset, it resulted in 6290, 6780, and 3542 crops for \textit{Suturing}, \textit{Needle-passing}, and \textit{Knot-tying}, respectively. All of these crops are new data samples for the network. The overall numbers of obtained crops are different since original recording lengths are varied across each trial in JIGSAWS. As a result, we obtained the total sample trials with the size of 6290, 6780, and 3542 for three tasks, respectively, according to the selected setting. 

\algnewcommand\algorithmicinput{\textbf{INPUT:}}
\algnewcommand\INPUT{\item[\algorithmicinput]}
\algnewcommand\algorithmicoutput{\textbf{OUTPUT:}}
\algnewcommand\OUTPUT{\item[\algorithmicoutput]}

\makeatletter
\newcommand{\removelatexerror}{\let\@latex@error\@gobble}
\makeatother

\begin{figure}[tb]
\removelatexerror
\begin{algorithm}[H]
\caption{Sliding-window Cropping Algorithm}
\label{alg: cropwindow}
\begin{algorithmic}[1]
\INPUT raw time series $X$, \textit{stepSize} $L$, \textit{windowWidth} $W$
\OUTPUT sub-sequences $s =$ SlidingWindow$(X, L, W)$
    \State \textbf{initialization} $‎m:= 0$, $n:=0$ 
    \State  $s: = empty$
  	\While {$m+W \leq \textit{length}(X)$}    
  		\State $s[n]:= X[m:(m+W-1)]$
        \State $m:= m + L, n:=n+1$
    \EndWhile
    \State  \Return sub-sequences $s$

\end{algorithmic}
 \end{algorithm}
\vspace{-0.8cm}
\end{figure}

\subsection{Training \& Testing}
To validate the model classification, we adopt two individual validation schemes in this work: \textit{Leave-one-supertrial-out (LOSO)} and \textit{Hold-out}. The objective of the comparison is to search for the best validation strategy suitable for system development in the case of deep learning. Based on each cross-validation setting, we train and test a surgical skill model for each surgical task, \textit{Suturing} (SU),  \textit{Knot-tying} (KT), and  \textit{Needle-passing} (NP).

\paragraph {Leave-one-supertrial-out (LOSO) cross-validation:} This technique involves repetitively leaving out a single subset for testing in multiple partitions. Specifically, a supertrial, $i$, defined as a subset of examples combining the $i$-th trials from all subjects for a given surgical task~\cite{gao2014JIGSAW}, is left out for testing, while the union of remaining examples is used for training. This process is repeated in five folds where each fold consists of each one of the five supertrials. The average of all five-fold performance measures (see Section 3.4 for definitions) in each test set is reported and gives an aggregated classification result. As a widely-used validation strategy, the \textit{LOSO} cross-validation shows its value in evaluating the robustness of a method for skill assessment. 

\paragraph {Hold-out:} Different from the \textit{LOSO} cross-validation, the \textit{Hold-out} strategy is implemented by conducting a train/test split once, which is normally adopted in deep learning models when large datasets are presented. In this work, one single subset consisting of one of the five trials from each surgeon, for a given surgical task, is left out throughout the training and used as a hold-out for the purpose of testing. Also, to reduce the bias and avoid potential overfitting, we randomly select a trial out of the five repetitions for each subject. 

\subsection{Modeling Performance Measures}
To compare the model performance, classifications are evaluated regarding four common metrics (Eq.~\ref{eq1})~\cite{sammut2011encyclopedia,kumar2012assessing,ahmidi2017dataset}: the average \textit{accuracy} -- ratio between the sum of correct predictions and the total number of predictions; \textit{precision} -- ratio of correct positive predictions ($T_p$) and the total positive results predicted by the classifier $(T_p + F_p)$;
 \textit{recall} -- ratio of positive predictions ($T_p$) and the total positive results in the ground-truth $(T_p + F_n)$; and \textit{f1-score} -- a weighted harmonic average between \textit{precision} and \textit{recall}.

\begin{equation}
\label{eq1}
\begin{aligned}
& \textit{precision} =  \frac{T_p}{T_p + F_p} \\[6pt]
& \textit{recall} = \frac{T_p}{T_p + F_n} \\[6pt]
& \textit{f1-score} =  \frac{2* (recall * precision) }{ recall + precision } \\[6pt]
\end{aligned}
\end{equation}
where $T_p$ and $F_p$ are the numbers of true positives and false positives,  $T_n$ and $F_n$ are the numbers of true negatives and false negatives, for a specific class.

In order to assess the computing effort involved in model classification, we measure the running time of skill models to classify all samples in the entire testing set. In the \textit{LOSO} scheme, the running time is measured as the average value from the five-fold cross-validation.

\begin{figure*}[tb]
      \centering
      \includegraphics [width=1\linewidth,clip,trim=0pt 5pt 5pt 10pt]{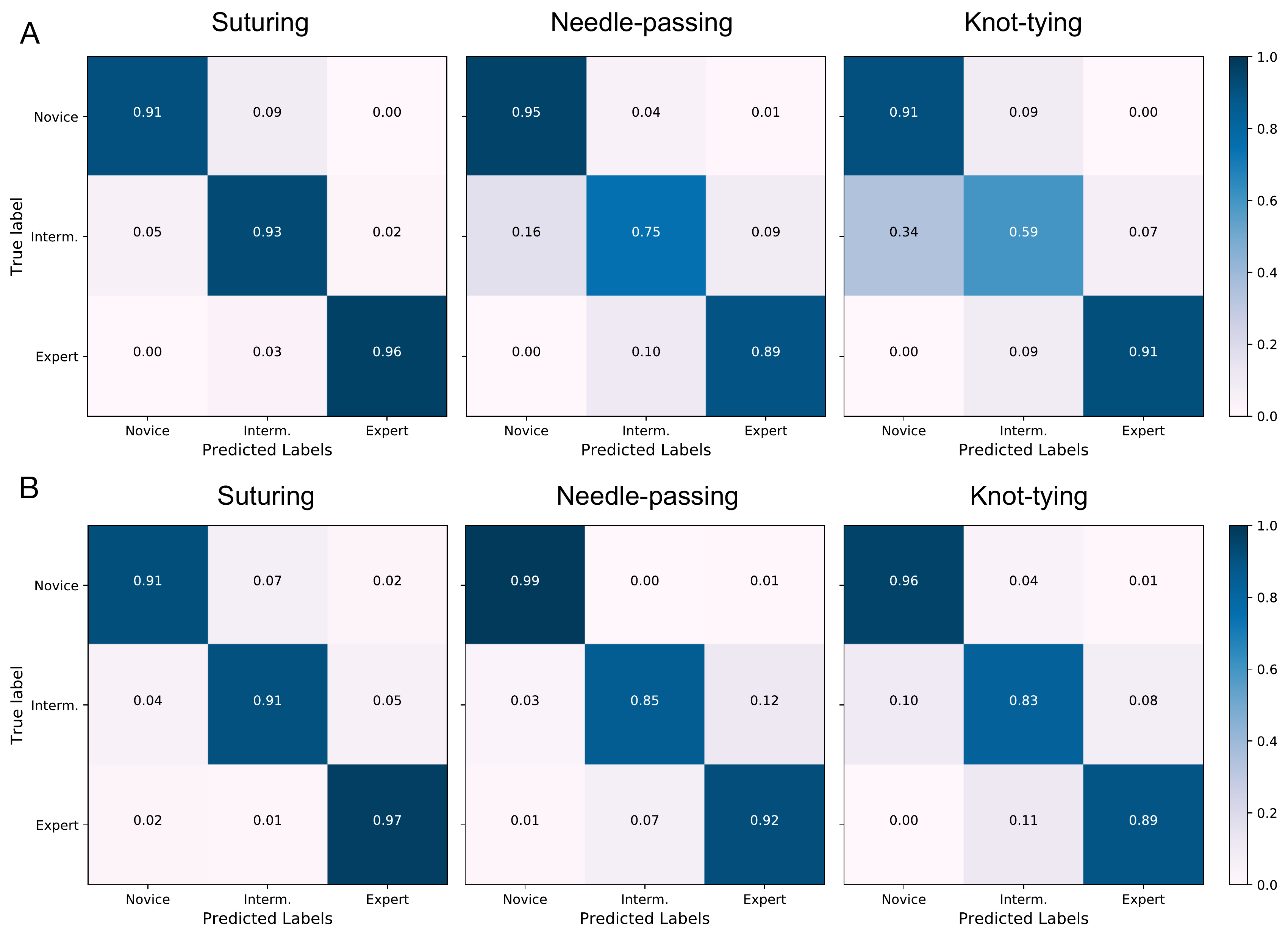}
      \caption{ {\bf Confusion matrices of classification results in three surgical training tasks. (A) self-proclaimed skill classification, (B) GRS-based skill classification.} Element value $(i,j)$ and color represent the probability of predicted skill label $j$, given the self-proclaimed skill label $i$, where $i$ and $j$ $ \in \{1 :``Novice", 2:``Intermediate", 3 :``Expert" \}$. The diagonal corresponds to correct predictions.}
      \label{fig: fig4}
\end{figure*}

\section{RESULTS}
\label{sec: results}
We evaluate the proposed deep learning approach for self-proclaimed skill classification and GRS-based skill classification using the JIGSAWS dataset.
The confusion matrices of classification results are obtained from the testing set under the \textit{LOSO} scheme.
We compare our results with the state-of-the-art classification approaches in Table~\ref{tab: review_perform}. 
It is important to mention that in order to obtain a valid benchmarking analysis, the classifiers investigated in this study are selected among the skill assessment using JIGSAWS motion data and evaluated based on the same \textit{LOSO} validation. 
Fig.~\ref{fig: fig4} (A) shows the results of three-class self-proclaimed skill classification. The proposed deep learning skill model achieved high-accuracy prediction performance. Specifically, our model obtained accuracies of 93.4\%, 89.8\%, and 84.9\% in tasks of \textit{Suturing}, \textit{Needle-passing} and \textit{Knot-tying}, respectively, using a window crop with 2-second duration containing 60 time steps ($W=60$). 
In contrast to the per-window assessment, highest accuracies reported in the literature range from 99.9\% to 100\% via a descriptive model using entropy features based on the entire observation of full operation trial. 
For the GRS-based skill classification, as shown in Fig.~\ref{fig: fig4} (B), the proposed approach can achieve higher accuracy than others (92.5\%, 95.4\%, and 91.3\% in \textit{Suturing}, \textit{Needle-passing} and \textit{Knot-tying}). Specifically, the deep learning model outperformed $k$-nearest neighbors (k-NN), logistic regression (LR), and support vector machine (SVM), with the accuracy improvements ranging from $2.89\%$ to $22.68\%$ in \textit{Suturing}, and $10.94\%$ to $21.09\%$ in \textit{Knot-tying}.

To study the capability of our proposed approach for online skill decoding, we further evaluate the performance of proposed approach using the input sequences with varying lengths. We repeated our experiment for the self-proclaimed skill classification with different sizes of sliding window: $W1=30$, $W2=60$ and $W3=90$.
Modeling performance of window sizes together with the average running time taken for self-proclaimed skill classification is reported in Table~\ref{tab: modelresults}.
The results show that our deep learning skill model can offer advantages over traditional approaches with highly time-efficient skill classification on the per-window basis, without the full observation of surgical motion for each trial (per-trial basis). 
Also, a higher average accuracy can be found with an increase of sliding window size. Specifically, the 3-second sliding window containing 90 time steps ($W3=90$) can obtain better results compared to 2-second window ($W2=60$), with average accuracy improvements of $0.75\%$ in \textit{Suturing}, $0.56\%$ in \textit{Needle-passing} and $2.38\%$ in \textit{Knot-tying}, respectively. 

Furthermore, in order to characterize the roles of two validation schemes, we repeat the above modeling process using \textit{Hold-out} strategy. Table~\ref{tab: modelresults} shows the comparison of self-proclaimed skill classification under \textit{LOSO} cross-validation and \textit{Hold-out} schemes.

\begin{table*}[tb]
\centering
\caption{ {\bf Comparison of existing algorithms employed for skill assessment using motion data from JIGSAWS dataset.} We benchmark the results in terms of accuracy based on \textit{LOSO} cross-validation. Models conducting classification on the trial level are categorized as \textit{per-trial basis}.  
}
\label{tab: review_perform}
\renewcommand\arraystretch{1.35}
\renewcommand\tabcolsep{11.5pt}
\begin{tabular}{lccccccc}
\toprule[\heavyrulewidth]         
 \multirow{2}{*}{\textbf{Author}}& \multirow{2}{*}{\centering \textbf{Algorithm}} & \multirow{2}{*}{ \parbox{1.5cm}{\center \textbf{Labeling\\Approach}}}  & \multirow{2}{*}{\parbox{1.5cm}{\centering \textbf{Metric\\ Extraction}}}  &  \multicolumn{3}{c}{\textbf{Accuracy}} &  \multicolumn{1}{c}{\multirow{2}{*}{\bf{Characteristics}} }\\ \cmidrule(lr){5-7}
 & & & & \textbf{SU} & \textbf{NP} & \textbf{KT} & \\ 
\midrule[\heavyrulewidth]      
 \multirow{2}{*}{\parbox{1.2cm}{Lingling\\\textit{2012}~\cite{tao2012sparse}}}   &   \multirow{2}{*}{S-HMM} & \multirow{2}{*}{\textit{Self-proclaim} }  & \multirow{2}{*}{ \parbox{1.5cm}{\centering{gesture\\segments}}} &  \multirow{2}{*}{97.4} &  \multirow{2}{*}{96.2} &  \multirow{2}{*}{94.4} & \multirow{2}{*}{ \parbox{3cm}{\textbullet\ generative modeling \\ \textbullet\ segment-based \\ \textbullet\ per-trial basis}}\\
 &  &  &  &  &\\
 \midrule[\heavyrulewidth]      
 \multirow{2}{*}{\parbox{1.2cm}{Forestier\\\textit{2017}~\cite{forestier2017jigsaw}}}  & \multirow{2}{*}{VSM}  & \multirow{2}{*}{\textit{Self-proclaim}} & \multirow{2}{*}{ \parbox{1.5cm}{\centering {bag of words\\features}}}   &  \multirow{2}{*}{89.7} &  \multirow{2}{*}{96.3} &  \multirow{2}{*}{61.1} & \multirow{2}{*}{ \parbox{3cm}{\textbullet\ descriptive modeling \\ \textbullet\ feature-based \\ \textbullet\ per-trial basis}}\\
 &  &  &  &  &\\
  \midrule[\heavyrulewidth]      
 \multirow{2}{*}{\parbox{1.2cm}{Zia \textit{2018}\\~\cite{zia2018skill}}}  & \multirow{2}{*}{NN}  & \multirow{2}{*}{\textit{Self-proclaim}}  & \multirow{2}{*}{ \parbox{1.5cm}{\centering{entropy\\features}}}  &  \multirow{2}{*}{100} &  \multirow{2}{*}{99.9} &  \multirow{2}{*}{100} & \multirow{2}{*}{ \parbox{3cm}{\textbullet\ descriptive modeling \\\textbullet\ feature-based \\ \textbullet\ per-trial basis}}\\
 &  &  &  &  &\\
 \midrule[\heavyrulewidth]     
 \multirow{3}{*}{\parbox{1.2cm}{Fard \textit{2017}\\~\cite{fard2018automated}}}   & $k$-NN  &  \multirow{3}{*}{\textit{GRS-based} }  &  \multirow{3}{*}{ \parbox{1.5cm}{\centering{movement\\features}}} & 89.7 & \textit{N/A} & 82.1 & \multirow{3}{*}{ \parbox{3cm}{ \textbullet\ descriptive modeling \\\textbullet\ feature-based\\ \textbullet\ two-class skill only \\ \textbullet\ per-trial basis}}\\
  & LR & & & 89.9 & \textit{N/A} & 82.3  & \\
  & SVM & &   & 75.4 & \textit{N/A} & 75.4 & \\
\midrule[\heavyrulewidth]     
 \multirow{3}{*}{\parbox{1.2cm}{\textbf{Current study}}}  & \multirow{3}{*}{ \centering CNN} & \multirow{1}{*}{\textit{Self-proclaim} } & \multirow{3}{*}{ \parbox{1.5cm}{\centering \textit{N/A}}}  & \multirow{1}{*}{ 93.4 } & \multirow{1}{*}{ 89.8} & \multirow{1}{*}{ 84.9 } & \multirow{1}{*}{ \parbox{3cm}{\textbullet\ deep learning modeling \\ \textbullet\ no manual feature \\\textbullet\ per-window basis \\ \textbullet\ online analysis}} \\ 
& &  \multirow{2}{*}{\textit{GRS-based}} &  & \multirow{2}{*}{ 92.5 } & \multirow{2}{*}{ 95.4} & \multirow{2}{*}{ 91.3 } & \\
 &  & &  &  &\\
\midrule[\heavyrulewidth]     
\end{tabular}
\end{table*}

\begin{table*}[tb]
\centering
\caption{ {\bf Summary table showing self-proclaimed skill classification performance based on different validation schemes and sliding windows.} Window size is set as $W1=30$, $W2=60$ and $W3=90$. Running time quantifies the computing effort involved in classification. Bold numbers denote best results regarding f1-score, accuracy, and running time.   
}
\label{tab: modelresults}
\renewcommand\arraystretch{0.86}
\renewcommand\tabcolsep{12pt}
\begin{tabular}{llccccccc}
\toprule[\heavyrulewidth]         
\multirow{2}{*}{\textbf{Task}} & \multirow{2}{*}{\parbox{1.0cm}{ \textbf{Validation Scheme}}} & \multirow{2}{*}{\parbox{0.8cm}{ \textbf{Window Size}}} & \multicolumn{3}{c}{\textbf{F1-score}} & \multirow{2}{*}{\textbf{Accuracy}}  & \multirow{2}{*}{\parbox{1.5cm}{\centering \textbf{Running Time ($ms$)}}}   \\ \cmidrule(lr){4-6}
 &  &  & \textbf{Novice} & \textbf{Interm.} & \textbf{Expert} &   & \\
\midrule[\heavyrulewidth]             
\multirow{6}{*}{\parbox{1cm}{\textbf{Suturing}}} & \multirow{3}{*}{LOSO} & $W1$ & 0.94 & 0.83 &  0.95 & 0.930  & \textbf{146.45}\\
 &  & $W2$ & 0.94 & 0.83 &  0.97 & 0.934  & 185.40\\
 &  & $W3$ &  \textbf{0.95} &  \textbf{0.86} & \textbf{0.96} &  \textbf{0.941}  & 247.01\\ \cmidrule(lr){2-8}
 & \multirow{3}{*}{Hold-out} & $W1$ &  0.98 & 0.92 &  0.94 &  0.961 & \textbf{98.10}\\
 &  & $W2$ & \textbf{0.99} & 0.94 & 0.96 & 0.972  & 146.40\\
 &  & $W3$ & \textbf{0.99} &  \textbf{0.98} & \textbf{0.97} & \textbf{0.983}  & 194.79\\
\midrule
\multirow{6}{*}{\parbox{1cm}{\textbf{Needle-\\passing}}} & \multirow{3}{*}{LOSO} & $W1$ &  0.95 & 0.73 & 0.88 & 0.889  & \textbf{153.36}\\
 &  & $W2$ &  0.95 & 0.75 & \textbf{0.90} & 0.898  & 194.98\\
 &  & $W3$ &  \textbf{0.96} & \textbf{0.76} & 0.89 & \textbf{0.903}  & 248.03\\ \cmidrule(lr){2-8}
 & \multirow{3}{*}{Hold-out} & $W1$ & 0.97 & 0.80 & 0.91 & 0.919  & \textbf{113.49}\\
 &  & $W2$ &  \textbf{0.98} &  0.81 & 0.91 &  0.925  & 169.72\\
 &  & $W3$ &  \textbf{0.98} &  \textbf{0.86} &  \textbf{0.94} & \textbf{0.945}  & 207.12 \\
 \midrule
\multirow{6}{*}{\parbox{1cm}{\textbf{Knot-tying}}} & \multirow{3}{*}{LOSO} & $W1$ & 0.90 & 0.57 & 0.90 & 0.847 & \textbf{101.83}\\
 &  & $W2$ & 0.90 & 0.62  & \textbf{0.92}  &  0.849  & 138.25\\
 &  & $W3$ &  \textbf{0.92} & \textbf{0.64} &  0.91 &  \textbf{0.868}  & 147.38 \\ \cmidrule(lr){2-8}
 & \multirow{3}{*}{Hold-out} & $W1$ &  0.87 &  0.42 & 0.91 & 0.803  & \textbf{74.5}\\
 &  & $W2$ & \textbf{0.88}  & \textbf{0.48} &  \textbf{0.92} & \textbf{0.817} & 113.55 \\
 &  & $W3$ &  \textbf{0.88} &  0.47 &  0.91 & 0.816  & 139.39\\
\bottomrule[\heavyrulewidth]     
\end{tabular}
\end{table*}

\section{DISCUSSION}
\label{sec: discussion}
Recent trends in robot-assisted surgery have promoted a great need for proficient approaches for objective skill assessment~\cite{vedula2017objective}. Although several analytical techniques have been developed, efficiently measuring surgical skills from complex surgical data still remains an open problem.
In this paper, our primary goal is to introduce and evaluate the applicability of a novel deep learning approach towards online surgical skill assessment. Compared to conventional approaches, our proposed deep learning model reduced dependency on the complex manual feature design or carefully-tuned gesture segmentation. Overall, deep learning skill models, with appropriate design choices, yielded competitive performance in both accuracy and time efficiency.

\subsection{Validity of our deep learning model for objective skill assessment}
For results shown in Fig.~\ref{fig: fig4} (A) and (B), we note that both \textit{Suturing} and \textit{Needle-passing} are associated with better results than \textit{Knot-tying} in both self-proclaimed skill classification and GRS-based skill classification, indicting that \textit{Knot-tying} is a more difficult task for assessment. For self-proclaimed skill classification, the majority of misclassification errors occurred during the \textit{Knot-tying} task where self-proclaimed $Intermediate$ are misclassified as actual $Novice$. As shown in Fig.~\ref{fig: fig4}(A), the distribution across $Intermediate$ is pronounced with the probability of $0.34$ being misclassified as $Novice$. 
%It indicates a slightly higher challenge to recognize $Intermediate$, which is likely to be confused with the $Novice$. 
%This mistake made by the network can be explained as there exist minor differences of skill levels between self-proclaimed $Intermediate$ and $Expert$ in the JIGSAW dataset~\cite{forestier2017jigsaw}. 
This could be attributed to the fact that the self-proclaimed skill labels, which are based on hours spent in robot operations, may not accurately reflect  the ground-truth knowledge of expertise. 
As evident, the classification using GRS-based skill labels generally performs better than the results using self-proclaimed skills. Our results indicate that more accurate surgeon skill labels relative to the true surgeon expertise might help to further improve the overall accuracy of skill assessment. 

As shown in Table~\ref{tab: review_perform}, high classification accuracy can be achieved by a few existing methods using generative modeling and descriptive modeling. Specifically, a generative model, sparse HMM (S-HMM), is able to give high predictive accuracy ranging from 94.4\% to 97.4\%. This result might benefit from a precise description of motion structures and pre-defined gestures in each task. However, such an approach requires prerequisite segmentation of motion sequences, as well as different complex class-specific models for each skill level~\cite{tao2012sparse}.  
Second, descriptive models sometimes may be superior to provide highly accurate results, such as the use of novel entropy features. However, the deficiency is that significant domain-specific knowledge and development is required to define the most informative features manually, which directly associate with the final assessment accuracy. This deficiency could also explain why there exists a larger variance in accuracy between other studies (61.1\%-100\%), which are sensitive to the choice of predefined features, as shown in Table~\ref{tab: review_perform}.
%In contrast, our method focuses on an automatic encoding of surgical skills via direct learning from raw motion data, and does not involve this time-consuming process. 

Another attention of our analysis is focused on the optimal sliding windows needed to render an efficient assessment. The duration of time steps in each window should roughly correspond to the minimum time required to decode skills from input signals. Usually, technical skill is assessed at the trial level; however, a quicker and more efficient acquisition may enable immediate feedback to the trainee, possibly improving learning outcomes.
Overall, our findings suggest that the per-window-based classification in this work is well-applicable for online settings. 
Smaller window size can allow for a faster running speed and less delay due to the light-weight computing expense. In contrast, an lager window size implies an increase of delay due to larger network complexity and higher computing effort involved in decoding. Specifically, as shown in Table~\ref{tab: modelresults}, within the \textit{LOSO} validation scheme, the network can classify the entire testing dataset within 133.88 $ms$ for $W1$ and 172.87 $ms$ running time for $W2$, while it required 214.14 $ms$ running time for $W3$ to classify the samples. 
However, it is important to mention that given an increase of window sizes, a higher accuracy can be achieved. In particular, there seems to be more gains in the \textit{Knot-tying} analysis, where the highest $2.24\%$ accuracy improvement was obtained from $W2$ to $W3$. This result might be due to the fact that more information of motion dynamics are contained in larger crops, thus allowing for an improved decoding accuracy.
We suggest that this trade-off between decoding accuracy and time efficiency could be a factor of interest in online settings of skill assessment.  

%%%%
\subsection{Comparison of Validation Schemes}
We investigated the validity of two different validation schemes for skill modeling. In this case, the differences between both are non-trivial in the deep learning development. 
Noticeably, \textit{LOSO} cross-validation gives a reliable estimate of system performance. However, the \textit{Hold-out} scheme, which uses a random subset of surgical trials as a hold-out, demonstrates relatively larger variances among results. This result can by explained by the differences among these randomly selected examples in the \textit{Hold-out} validation.
Nevertheless, the \textit{Hold-out} shows consistency with the results in \textit{LOSO} scheme across different tasks and window sizes, as shown in Table~\ref{tab: modelresults}.
It is important to note that given a large dataset, the \textit{LOSO} cross-validation might be less efficient for model assessment. In this scenario, the computing load in \textit{LOSO} modeling has been largely increased, which may not be suitable for complex deep architectures. However, the \textit{Hold-out} only needs to run once and is less computationally expensive in modeling.

\subsection{Limitations}
Despite the progress in present work, there still exist some limitations of deep learning models towards a proficient online skill assessment.
First, as confirmed by our results, the classification accuracy of supervised deep learning relies heavily on the labeled samples. The primary concern in this study lies with the JIGSAWS dataset and the lack of strict ground-truth labels of skill levels. It is important to mention that there is a lack of consensus in the ground-truth annotation of surgical skills. In the GRS-based labeling, skill labels were annotated based on the predefined cut-off threshold of GRS scores, however, no commonly accepted cutoff exists. 
For future work, a refined labeling approach with stronger ground-truth knowledge of surgeon expertise may further improve the overall skill assessment~\cite{sun2017revisiting,dockter2017minimally}.
Second, we will search for a detailed optimization of our deep architecture, parameter settings and augmentation strategies to better handle motion time-series data and improve the online performance further. 
In addition, the interpretability of automatically learned representations is currently limited due to the black-box nature of deep learning models. It would be interesting to investigate a visualization of deep hierarchical representations to understand hidden skill patterns, so as to better justify the decision taken by a deep learning classifier.

\section{CONCLUSION}
\label{sec:conclusion}
 The primary contributions of this study are: (1) a novel data-driven deep architecture for an active classification of surgical skill via end-to-end learnings, (2) an insight in accuracy and time efficiency improvements for online skill assessment, and (3) application of data augmentation and exploration of validation schemes feasible for deep skill modeling.  
Taking advantage of recent technique advances, our approach has several desirable proprieties and is extremely valuable for online skill assessment.
First, a key benefit is an end-to-end skill decoding, learning abstract representations of surgery motion data with automatic recognitions of skill levels. Without a priori dependency on engineered features or segmentation, the proposed model achieved comparable results to previously reported methods. 
It yielded highly competitive time efficiency given relatively small crops (1$-$3 second window with 30$-$90 time steps), which were computationally feasible for online assessment and immediate feedback in training.
Furthermore, we demonstrated that an improvement of modeling performance could be achieved by the optimization of design choices. An appropriate window size could provide better results in \textit{Knot-tying} with a 2.24\% accuracy increase. Also, the development of deep skill models might benefit from the \textit{Hold-out} strategy, which requires less computing effort than the \textit{LOSO} cross-validation, especially in the case where large datasets are involved.  

Overall, the ability to automatically learn abstract representations from raw sensory data with high predictive accuracy and fast processing speed, makes our approach well-suited for online objective skill assessment. 
The proposed deep model can be easily integrated into the pipeline of robot-assisted surgical systems and could allow for immediate feedback in personalized surgical training. 

\section*{Acknowledgment}
This work is supported by National Science Foundation (NSF\#1464432).
\section*{Conflict of interest }
The authors, Ziheng Wang and Ann Majewicz Fey, declare that they have no conflict of interest.
\section*{Ethical approval}
For this type of study formal consent is not required.
\section*{Informed consent}
This articles does not contain patient data.

%%%%%%%%%%%%%%%%%%%%%%%%%%%%%%%%%%%%%%%%%%%%%%%%%%%%%%%%%%%%%%%%%%%%%%%%%%%%%%%%
\bibliographystyle{spbasic.bst}
\bibliography{citation_DL.bib}

%%%%%%%%%%%%%%%%%%%%%%%%%%%%%%%%%%%%%%%%%%%%%%%%%%%%%%%%%%%%%%%%%%%%%%%%%%%%%%%%
\end{document}